\documentclass[letterpaper, 10 pt, conference]{ieeeconf}

\usepackage{epsfig}
\usepackage{color,booktabs}
\usepackage{graphicx}
\usepackage[figuresright]{rotating}
\usepackage{amssymb}
\usepackage{amsmath}
\usepackage{setspace}
\usepackage{rotating}
\usepackage{multirow}
\usepackage{wrapfig}
\usepackage{booktabs}
\usepackage{array}
\usepackage{comment}
\usepackage{pict2e}
\usepackage{epstopdf}
\usepackage[ruled]{algorithm2e}
\usepackage{xcolor}

\usepackage{balance}
\usepackage{mathtools}
\usepackage{url}  %
\usepackage{subcaption}
\usepackage[font=footnotesize]{caption}
\usepackage{xparse}
\usepackage[T1]{fontenc} %

\usepackage{colortbl}

\NewDocumentCommand\jj{+u{\jj}}{\ignorespaces}

\begin{document}

\title{\LARGE \bf Deep Multi-Task Learning for Anomalous Driving Detection Using CAN Bus Scalar Sensor Data}
\author{{\bf Vidyasagar Sadhu, Teruhisa Misu*, Dario Pompili}\\
Department of ECE, Rutgers University--New Brunswick, NJ; *Honda Research Institute, Mountain View, CA\\
\textit{\{hss64, pompili\}@rutgers.edu; *tmisu@honda-ri.com}
}

\maketitle

\thispagestyle{empty}
\pagestyle{plain} \pagenumbering{arabic}

\begin{abstract}
Corner cases are the main bottlenecks when applying Artificial Intelligence~(AI) systems to safety-critical applications. An AI system should be intelligent enough to detect such situations so that system developers can prepare for subsequent planning.
In this paper, we propose semi-supervised anomaly detection considering the imbalance of normal situations. In particular, driving data consists of multiple positive/normal situations (e.g., right turn, going straight), some of which (e.g., U-turn) could be as rare as anomalous situations. Existing machine learning based anomaly detection approaches do not fare sufficiently well when applied to such imbalanced data. In this paper, we present a novel multi-task learning based approach that leverages domain-knowledge (maneuver labels) for anomaly detection in driving data. We evaluate the proposed approach both quantitatively and qualitatively on 150 hours of real-world driving data and show improved performance over baseline approaches. 

\end{abstract}

\section{Introduction}\label{ref:intro}

\textbf{Overview:} 
Autonomous Driving~(AD) aka self-driving vehicles are cars or trucks in which human drivers are never required to take control to safely operate the vehicle. They combine sensors and software to control, navigate, and drive the vehicle. Though still in its infancy, self-driving technology is becoming increasingly common and could radically transform our transportation system economy and society. Many thousands of people die in motor vehicle crashes every year in the United States (more than 30,000 in 2015); self-driving vehicles could, hypothetically, reduce that number---software could prove to be less error-prone than humans. Based on automaker and technology company estimates, level~4 (the car is fully-autonomous in some driving scenarios, though not all) self-driving cars could be for sale in the next several years~\cite{level4volvo}. 

\textbf{Motivation:}
Currently, safety is an overarching concern in AD technology, that is preventing its full deployment in real world. The question to answer is whether the AD system can handle potentially dangerous and anomalous driving situations. Before AD systems can be fully deployed, they need to know how to handle such scenarios, which in turn calls for heavy training of an AD system on such scenarios. The challenge lies in that, these scenarios are very rare. They constitute the `long tail of rare events' (on a Gaussian considering all events) and comprise less than $0.01\%$ of all events in a given driving dataset. Hence Artificial Intelligence~(AI) technology need to be used to both mine these ``gems'' in a given dataset and then to train the AD system to handle such `special' situations. 
Detecting such anomalous driving scenarios that is crucial to building fail-safe AD systems offers two advantages---both offline and online. In the former, given a dataset, the identified anomalous driving scenarios can be used to train an AD system to better handle such scenarios. This can be achieved, for example, via weighted training---give more weight to learning anomalous scenarios than normal scenarios. 
For online purposes, detecting anomalous driving scenarios ahead of time can help prevent accidents in some cases, by taking a corrective action so as to steer the system in a safe direction (e.g., apply appropriate control signals or if possible, handing over the control to a human driver). 
We specifically consider only the Controller Area Network~(CAN) bus sensor data such as pedal pressure, steer angle, etc. (multi-modal time-series data) due to its simplicity, while still providing valuable (though not complete) information about the driving profile; augmenting with video data will be our future work. We consider any unusual pattern (such as abruptness, rarity, etc.) among different modalities as a sign of an anomaly. Such a pattern could have happened due to unusual reaction of driver on the pedal, accelerator, steering wheel, etc. which in turn implies the driver has gone through a challenging (anomalous) driving situation. Though model-based (rule-based) approaches can be used to detect anomalies in multi-modal time-series data, 
they are good only for simple cases such as threshold based anomaly detection (speed/deceleration greater than a threshold, etc.). It is difficult as well as tedious to compose rules for complex and even unknown (apriori) situations. On the other hand, data-driven approaches, can learn representations directly from the data, and use them to detect anomalies. This gives them the ability to detect complex and unknown anomalies directly from data. Though, the performance of data-driven approaches is only as good as the data, this limitation can be addressed to some extent, when large amount of data is considered for training. In data-driven approaches, deep-learning based approaches (as opposed to classical machine learning techniques) are especially interesting due to their ability to learn the features on their own without the need for expert domain-expertise. Existing deep learning approaches for anomaly detection in multi-modal time series data include reconstruction-error based approaches such as Long Short Term Memory~(LSTM) autoencoders. These approaches do not perform well in the case where multiple ``normal'' situations (multiple positive classes) exist with class imbalance problem. In case of driving data, these classes correspond to right-turn, going-straight, u-turn, etc. where the data for u-turn is far lesser than the data for going-straight. There is a higher chance that the classifier in these approaches overfits smaller (less-frequent) classes resulting in poor performance. Since reconstruction error is used as a measure of anomaly by these approaches, they classify the less frequent normal classes (e.g., u-turn) also as anomalous, further degrading the performance. 

\begin{figure}
\begin{center}
\includegraphics[width=3.6in]{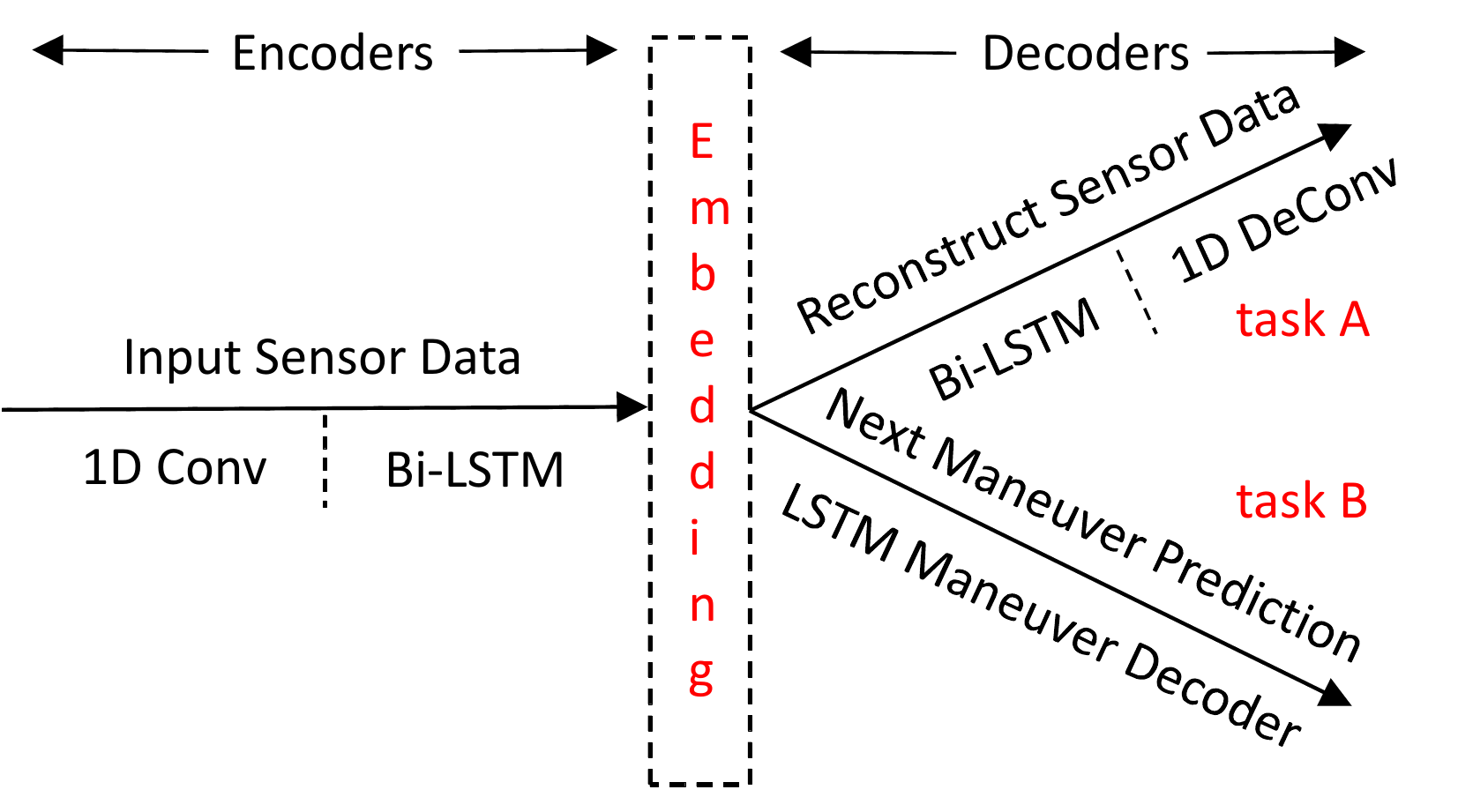}
\end{center}
\caption{A high-level overview of the proposed multi-task deep learning based approach for anomaly detection in multi-modal time-series driving dataset.}\label{fig:lstm_multitask_highlevel}
\end{figure}

\textbf{Our Approach:}
We make the observation that, while reconstruction-error based approaches perform poorly with rare but non-anomalous events, their performance can be greatly improved with the help of simple domain-knowledge (availability of maneuver labels in our case). Leveraging these maneuver labels, we add a symbol predictor to the autoencoder system (creating a multi-task learning system) which acts as a regularizer to the autoencoder, thereby achieving better performance than a standalone autoencoder system.
The proposed deep multi-task learning based anomaly detection system is shown in Fig.~\ref{fig:lstm_multitask_highlevel}. The two tasks in the proposed approach are a convolutional bi-directional LSTM~(Bi-LSTM) based autoencoder and a convolutional Bi-LSTM based sequence-to-sequence~(seq2seq) symbol predictor (in contrast to simple LSTM predictor that predicts raw sensor data rather than symbols). In the seq2seq predictor, the predicted symbols/labels correspond to automobile's next series of maneuvers (e.g., going-straight, left-turn, etc.). These labels are obtained from manually annotated driving data. We show that the proposed multi-task learning approach performs better than existing deep learning based anomaly detection approaches such as LSTM autoencoder and LSTM predictor as one task acts as a regularizer for the other. In addition to reconstructing the input data (via autoencoder), the network is also constrained to predict the next series of maneuvers (via symbol predictor) and as such the chance of overfitting is reduced. Such a regularizer system also helps solve the problem of overfitting to smaller class mentioned above. 
Secondly, 
the proposed multi-task learning approach leverages these maneuver labels to define a custom anomaly metric (rather than simple reconstruction error) that weighs down detection of rare but non-anomalous patterns such as u-turns as anomalies. The approach has been tested on 150 hours 
of raw driving data~\cite{ramanishka2018CVPR} in and around Mountain View, California, USA and is shown to perform better than state-of-the-art approach, LSTM-based autoencoder~\cite{Malhotra2016}.

\textbf{Our contributions} can be summarized as follows.
\begin{itemize}
\item We propose a novel multi-task learning (convolutional BiLSTM autoencoder and symbol predictor) approach for detecting anomalous driving with multiple ``normal'' classes and a class imbalance problem. Our approach leverages simple domain-knowledge (manuever labels) to build a regularizer system that reduces overfitting and enhances overall reconstruction performance.

\item We propose an anomaly scoring metric that leverages such maneuver labels and reduces the cases where rare, but non-anomalous, events are classified as anomalies.

\item We evaluated our approach both quantitatively and qualitatively on 150 hours of real driving data and compare it with state-of-the-art LSTM autoencoder and multi-class LSTM autoencoder approaches to show its advantages over them.
\end{itemize}

\textbf{Paper Outline:} %
In Sect.~\ref{sec:rel_work}, we position our work with respect to state of the art; in Sect.~\ref{sec:prop_soln}, we present our approach, and in Sect.~\ref{sec:perf_eval}, we evaluate it both quantitatively and qualitatively on real driving data. Finally, in Sect.~\ref{sec:conc}, we conclude the paper and provide future directions.

\section{Related Work}\label{sec:rel_work}

In this section, we describe important related work in the domain of anomaly detection for multi-modal/multi-variate time series data. Anomaly detection is generally an unsupervised machine learning (ML) technique due to lack of sufficient examples for the anomalous class. Within unsupervised learning, it can be broadly classified into the following categories---contextual anomaly detection, ensemble based methods and finally deep learning approaches. These methods internally use statistical/regression based approaches, dimensionality reduction, distribution based approaches. In statistical approaches, features are generally hand-made from the data such as mean, variance, entropy, energy, etc. Certain statistical tests/formal rule checking actions are performed on these features to determine if the data is anomalous. In dimensionality reduction, the data is projected onto a low-dimensional representation (such as princial components in Principal Component Analysis~(PCA)). The idea is that, this low-dimensional representation captures the most important features of the input data. Then clustering techniques such as k-means or Gaussian Mixture Models~(GMMs) are used to cluster these low-dimensional features to identify anomalies. In distribution-based approaches, the training data is fit to a distribution (such as multi-variate gaussian distribution or a mixture of them). Then given a test point, distance is calculated of this test point from the fitted distribution (e.g., using Mahalanobis distance) that represents the measure of anomaly.

\textbf{Contextual anomaly detection:} An anomaly may not be considered anomaly when the context under which it happens is well-known. For example, the CANbus sensor data of a car may look anomalous when the car is taking a U-turn, which is not considered an anomaly. This is also called seasonal anomaly detection in other domains such as building energy consumption, retail sales, etc. Hayes et al.~\cite{Hayes2015ContextualData} and Capozzoli et al.~\cite{Capozzoli2015FaultBuildings} present a two-step approach for contextual anomaly detection. In the former, in step~1, only the sensor's past data is used to identify anomalous behavior. For this it uses univariate Gaussian function. 
Later in step 2, if the output of step 1 is found to be anomalous, then it passes to step 2 to check if it is contextually anomalous or not. 
Twitter~\cite{twitter_ad} recently published a seasonal anomaly detection framework based on Seasonal Hybrid Extreme Studentized Deviate test (S-H-ESD). Netflix~\cite{netflix_ad} too recently released an approach for anomaly detection in big data using Robust PCA~(RPCA). Even though Netflix's approach seemed successful, their statistical approach relies on high dimensionality datasets to compute a low rank approximation which limits its applicability. Finally, Toledano et al.~\cite{Toledano2018} propose a bank filter and fast autocorrelation based approach to detect anomalies in large scale time-series data considering seasonal variations.

\textbf{Ensemble based methods:} In ensemble learning, different models are trained on the same data (or random sets of samples from the original data) and a majority voting (or another fusion technique) is used to decide the final output. Another advantage of ensemble learning is that the member models are chosen such that they are complementary to each other in terms of their strengths/weaknesses, i.e., the weaknesses of one are compensated by the strengths of the other. For example, Araya et al.~\cite{Araya2017AnConsumption}, proposed an ensemble based collective and contextual anomaly detection framework. The ensemble consisted of pattern recognition algorithms such as Autoencoder and PCA, as well as prediction based anomaly detectors such as Support Vector Regression~(SVR) and Random Forest. They showed that the ensemble classifier is able to perform well compared to the base classifiers.

\textbf{Deep learning methods:} In deep learning techniques, the features are generally learned by the classifier itself, so there is no need to hand-engineer these features. The techniques within this can be broadly classified into two categories: 
(i) Representation learning for reconstruction: Here the input data is mapped to a latent space (generally lower dimension than input data) using an encoder and then the latent space is remapped to input space using a decoder. The latent space captures a representation of the input data similar to PCA. The reconstruction error at the end of this process is a measure of anomaly. Autoencoders are prime examples in this category. For example, Malhotra et al.~\cite{Malhotra2016} present an LSTM based encoder-decoder approach for multi-sensor time-series anomaly detection. The approach has been tested on multiple datasets including power demand, space shuttle valve, medical cardiac data and a proprietary engine data and showed promising results.
(ii) Predictive modeling: Here the current/future data is predicted from past data using LSTM modules that capture long term temporal behavior. The prediction error is a measure of anomaly. LSTM sequence predictors are examples in this category. For example, Taylor et al.~\cite{Taylor2016AnomalyNetworks} proposed an LSTM predictor based anomaly detection framework for automobiles based on Controller Area Network~(CAN) bus data of an automobile similar to ours. 
Hallac et al.~\cite{Hallac2018Drive2Vec:Data} present an embedding approach for driving data called Drive2vec which can be used to encode the identity of the driver. However this approach only complements ours, as our approach can work both with raw data as well as embedded data. Malhotra et al.~\cite{Malhotra2015LongSeries} proposed an LSTM based predictor for anomaly detection in time series data that is shown to perform well on four kinds of datasets mentioned above.

In contrast to these approaches, we propose a multi-task deep learning based approach, that overcomes the shortcomings on (i) and (ii) by--- incorporating a built-in regularizer (as one task acts as regularizer for the other) and leveraging domain knowledge (such that rare but non-anomalous maneuvers such as U-turns are not classified as anomalies).

\section{Proposed Solution}\label{sec:prop_soln}
In this section, we first explain the LSTM autoencoder (reconstruction-error) based approach~\cite{Malhotra2016} which is currently the best performing (unsupervised) anomaly detection framework for multi-modal time-series data. We then present our semi-supervised approach for anomaly detection in driving data which leverages the maneuver labels to improve the performance. 
Anomaly detection using unsupervised learning consists of two steps. In step~1, the system is trained with several normal examples to learn representations of the input data e.g., GMM clustering. Because we are dealing with temporal data, a sliding window approach 
needs to be adopted to learn these representations. In step~2, given a test data point, we define an anomaly score based on the learned representations, e.g., distance from the mean of the cluster.

\subsection{LSTM Autoencoder (Existing Approach)}\label{sec:prop_soln:lstm_ae}

\begin{figure}
\begin{center}
\includegraphics[width=3.4in]{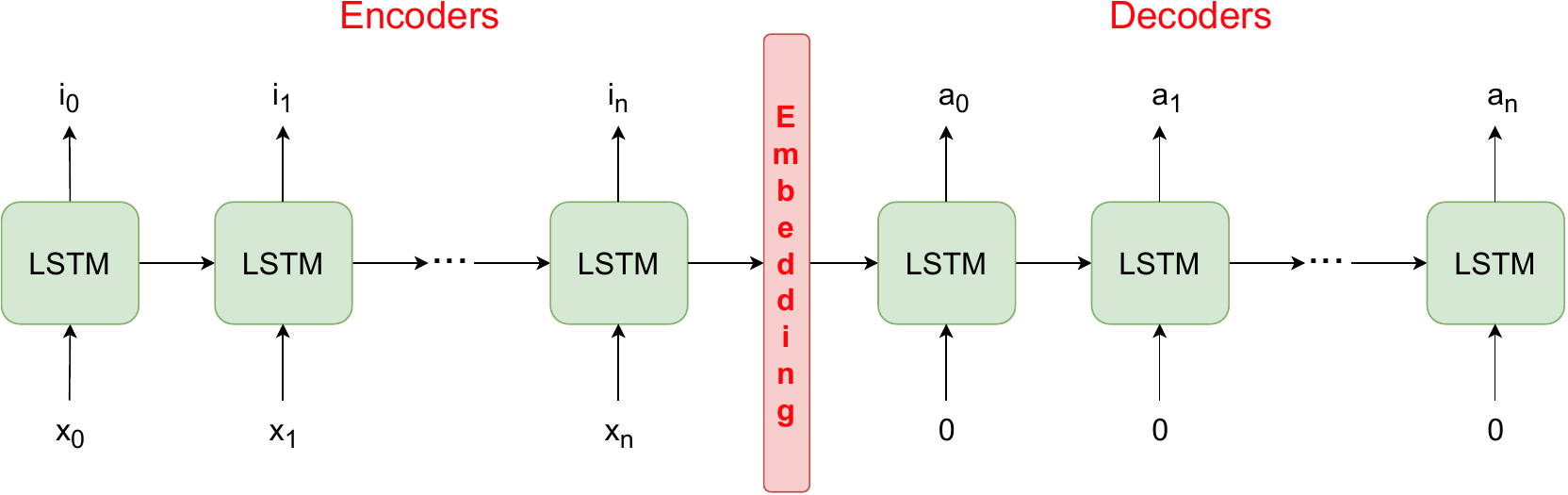}
\end{center}
\caption{LSTM Autoencoder---the encoder cells encode the input data into a representation that is stored in the cell state of the last encoder LSTM cell. The decoder cells take it as input and try to generate the time series data.}\label{fig:lstm_ae}
\end{figure}

Fig.~\ref{fig:lstm_ae} shows the LSTM autoencoder high-level architecture. Input time series data, $\{x_0, x_1,...x_n\}$, of size $n+1$ (corresponding to one window of data segmented from full data) is fed to the encoders which consist of $n+1$ LSTM cells. Each LSTM cell encodes its input and the cell state from previous cell into its own cell state, which is passed onto the next LSTM cell. Finally, the cell state of the last LSTM cell has the encoded representation---which we call \textit{embedding}---of all the input data $\mathbf{x}$. The size of this embedding is equal to the number of units (also called hidden size) in the last LSTM cell. The decoders similarly consist of a series of LSTM cells, however the input to these decoders is given as zero as the goal is to regenerate the input data. Another approach of feeding the output of the previous cell ($a_i$) as input to the next cell is also possible. The first decoder LSTM cell takes the embedding as one of the inputs (the other input being zero) and passes on its cell state to the next decoder cell. The process is repeated for $n+1$ time steps. During each step, the LSTM cell generates an output $a_i$, finally resulting in $\{a_0, a_1,...a_n\}$ after $n+1$ steps. The network is trained by minimizing the difference, $|\mathbf{x} - \mathbf{a}|^2$ using stochastic gradient descent and backpropagation. After sufficient training, the network is able to learn good representations of the input data stored in its embedding, which completes step~1. The network is then able to reconstruct new data very well i.e., with lower reconstruction error, as long as it has seen similar pattern data during training. However, when the network is fed with data that has completely different pattern than is used during training, there will be a large reconstruction error.

\begin{figure}
\begin{center}
\includegraphics[width=3.4in]{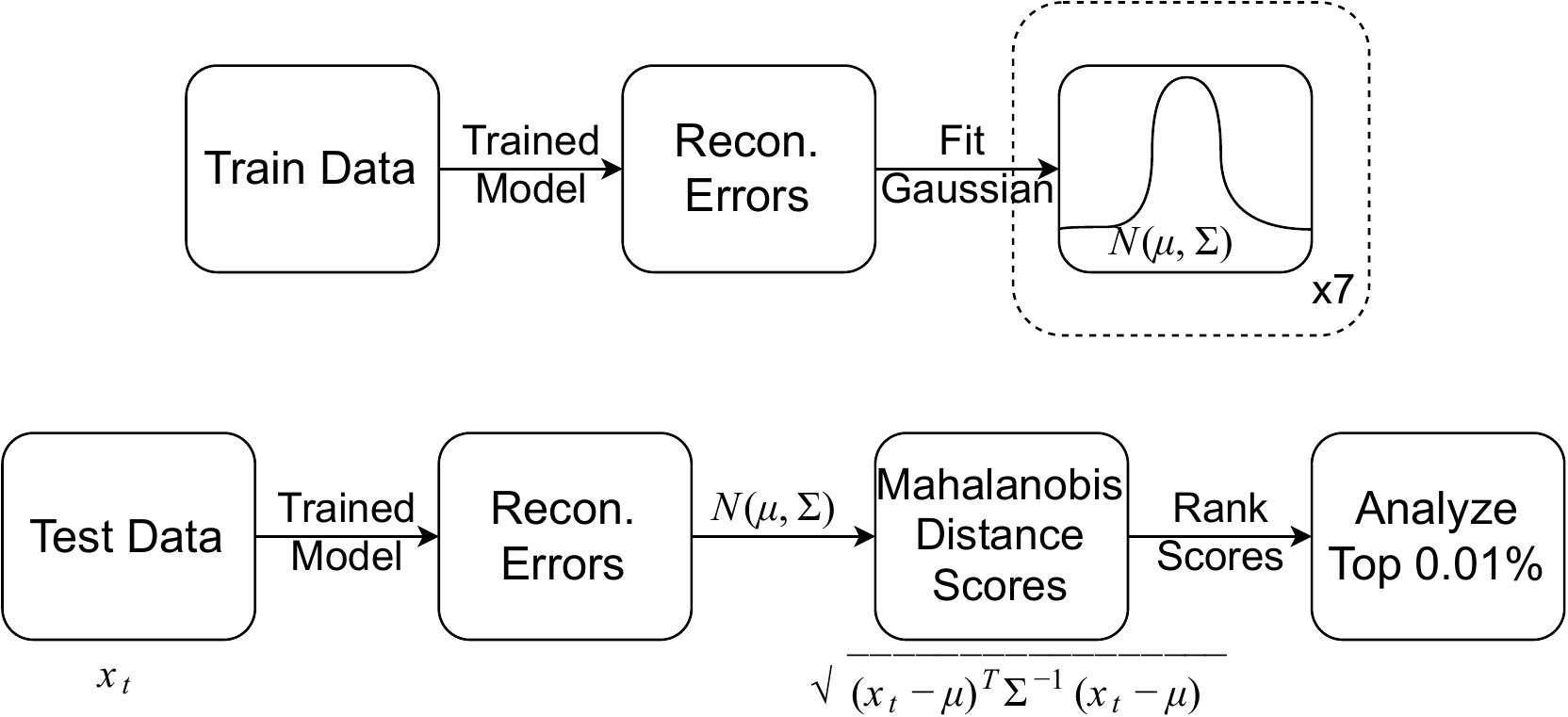}
\end{center}
\caption{(Top) After the model is trained, we fit the reconstruction errors to a multivariate gaussian model. (Bottom) Given a test data point, we first find the reconstruction error and then find the mahalanobis distance/score of this point with respect to the fitted gaussian distribution. Top scores can then be analyzed as per requirements.}\label{fig:anom_scores_combined}
\end{figure}

Though reconstruction error can directly be used as a measure of anomaly for step~2, better results can be achieved, with further processing. The method currently adopted~\cite{Malhotra2016} is shown in Fig.~\ref{fig:anom_scores_combined}. After the network is trained, the train data is again fed to the trained network to capture the reconstruction errors. These errors are then fit to a multivariate gaussian distribution as shown in Fig.~\ref{fig:anom_scores_combined}. Given a test data point, the reconstruction error is first calculated using the trained model. Mahalanobis distance of the error is then calculated with respect to the fitted gaussian model using the formula shown in Fig.~\ref{fig:anom_scores_combined}. These distances, which are considered anomaly scores are then sorted in decreasing order and analyzed as per requirements e.g., analyze top 0.01\%.

\subsection{Multi-task Learning (Proposed Approach) }\label{sec:prop_soln:multi_task}
As mentioned earlier, fully unsupervised reconstruction error based approaches such as LSTM autoencoder fare poorly when there are rare occurring positive (non-anomalous) classes in the data. For such relatively rare cases, the network is unable to learn representations, thereby producing large reconstruction error. We solve this problem, by designing a semi-supervised multi-task learning framework that leverages driving maneuver labels as shown in Fig.~\ref{fig:lstm_multitask_highlevel}. Here task~A is the autoencoder, while task~B is a symbol/maneuver predictor. Task~B acts as a regularizer to the autoencoder as the overall network is also constrained to predict the next series of maneuvers apart from reconstructing the input data. For this to be possible, better representations need to learned by the network that can help in both reconstruction and prediction. This combined (multi-task) system performs better reconstruction than a standalone autoencoder. Similarly, autoencoder system (task~A) acts as a regularizer for symbol predictor (task~B). This is because autoencoder helps in learning good representations of the input data. These representations can then be used by the symbol predictor to predict next series of maneuvers. Hence, in a similar way as mentioned above, the combined system produces a better symbol predictor than a standalone symbol predictor. Both these are possible as both tasks mutually help each other. Our approach is semi-supervised as we make use of maneuver labels to design a regularizer in Task~B, but is not supervised as we do not have anomaly and non-anomaly labels. We will now explain the encoder and the decoders of both tasks in detail.

\begin{figure}
\begin{center}
\includegraphics[width=3.4in]{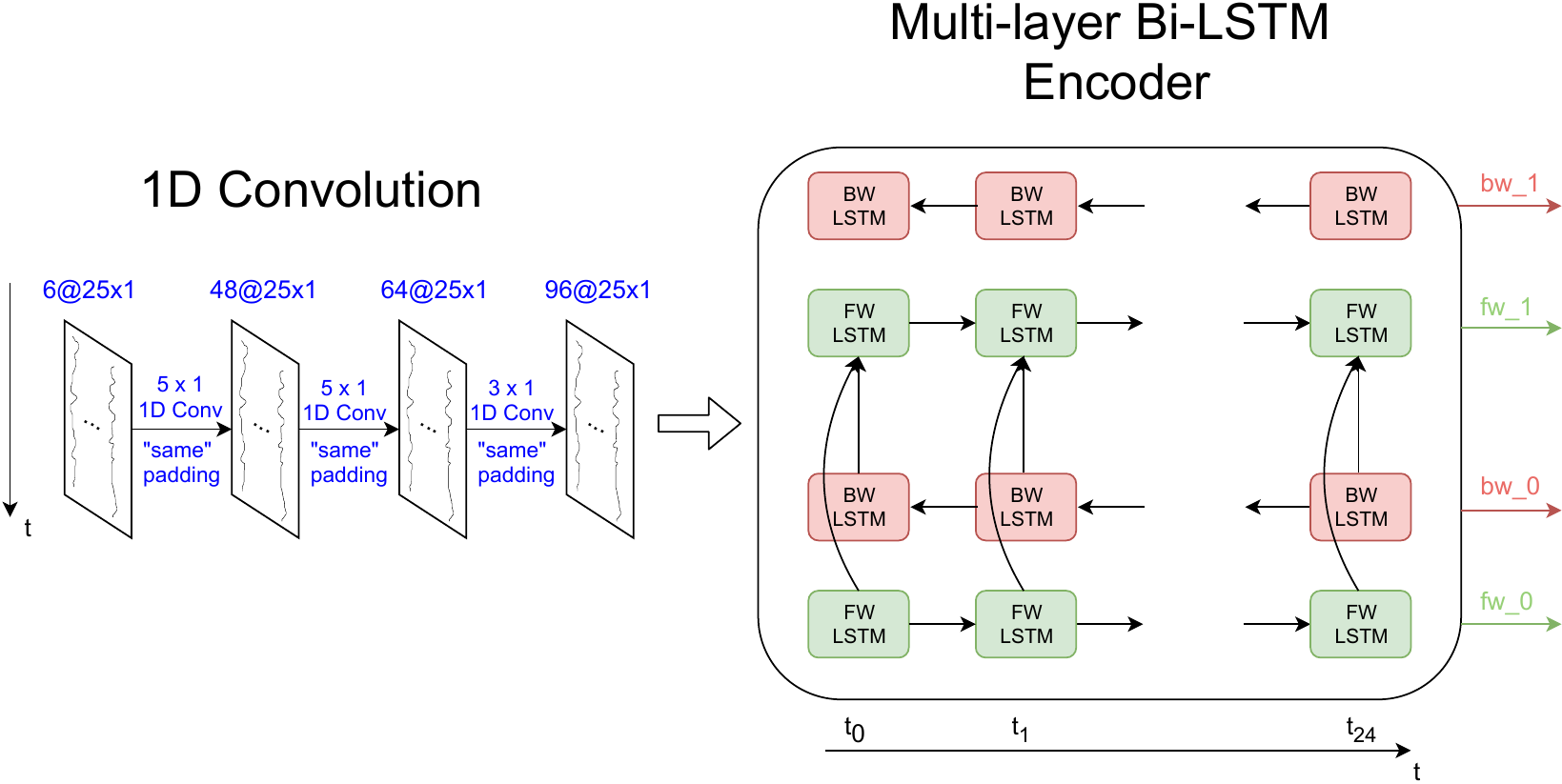}
\end{center}
\caption{Convolutional and Bi-LSTM encoder of the proposed multi-tasking learning framework in  Fig.~\ref{fig:lstm_ae}.}\label{fig:multi_task_lstm_encoder}
\end{figure}

\textbf{Convolutional Bi-LSTM Encoder:} The basic encoder in an LSTM autoencoder (Fig.~\ref{fig:lstm_ae}) does not perform sufficiently well as it not does take into account: (i) inter channel correlations (ii) directionality of data. We design an encoder that addresses these issues as shown in Fig.~\ref{fig:multi_task_lstm_encoder}. It consists of a series of 1-dimensional~(1D) convolutional layers followed by bi-directional LSTM layers. The convolutional layers help in capturing inter-channel spatial correlations, while the LSTM layers help in capturing inter- and intra-channel temporal correlations. Unidirectional LSTM layers capture temporal patterns only in one-direction, while the data might exhibit interesting patterns in both directions. Hence to capture these patterns, we have a second set of LSTM cells for which the data is fed in the reverse order. Further, we have multiple layers of these bi-directional LSTM (bi-LSTM) layers to extract more hierarchical information. All the data that has been processed through multiple convolutional and bi-LSTM layers is available in the cell states of final LSTM cells. This is the output of the encoder which will be fed as input to the decoder tasks.

\begin{figure}
\begin{center}
\includegraphics[width=3.4in]{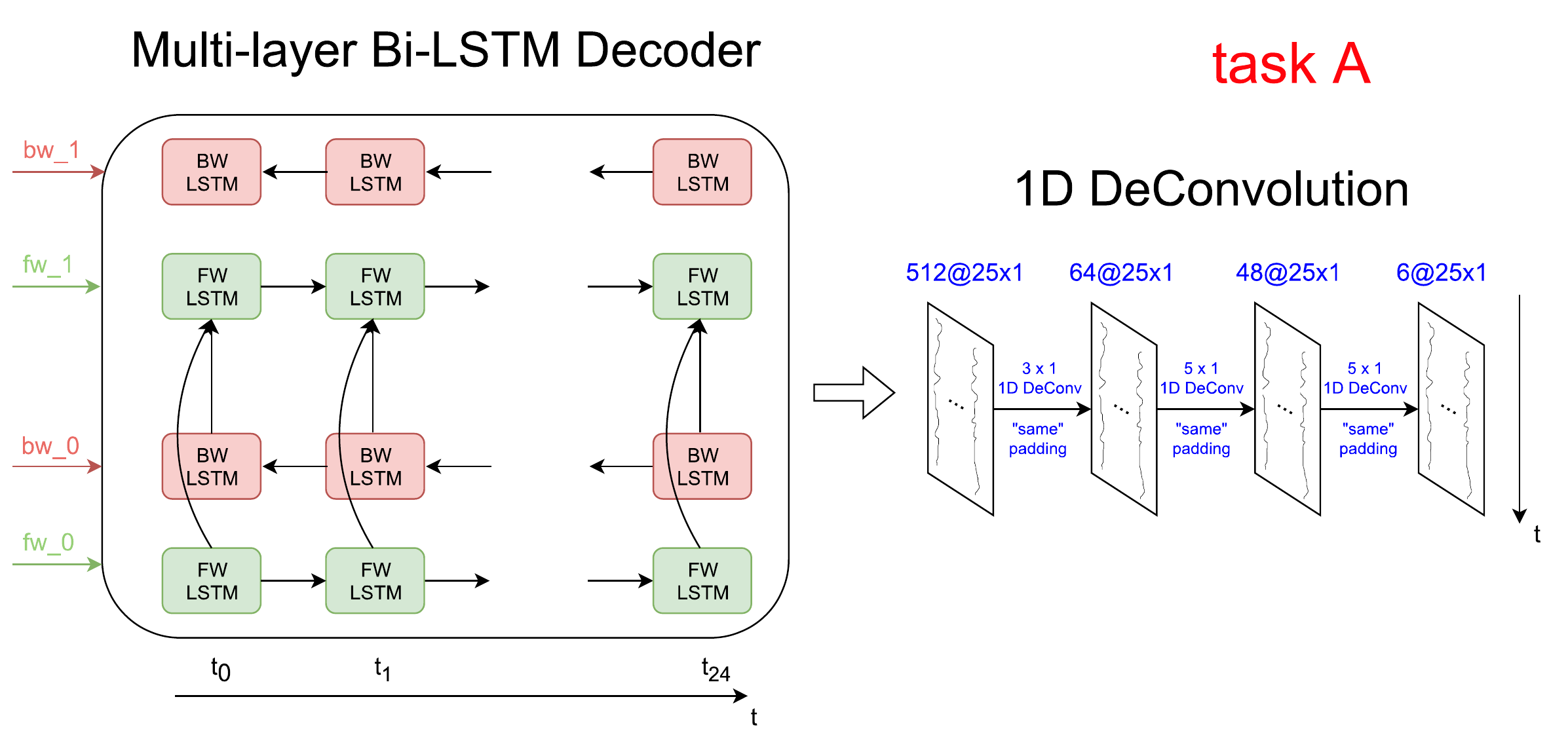}
\end{center}
\caption{Convolutional Bi-LSTM decoder for task~A (autoencoder) in Fig.~\ref{fig:lstm_multitask_highlevel}.}\label{fig:lstm_multitask_decoder_taskA}
\end{figure}

\textbf{Decoder (Autoencoder, Task~A)}: The decoder in autoencoder (task~A) performs encoder operations in reverse order so as to reconstruct the input data (Fig.~\ref{fig:lstm_multitask_decoder_taskA}). It first consists of bi-LSTM layers which take the final cell states from encoder as one of the inputs (the other input being zero). As mentioned in Sect.~\ref{sec:prop_soln:lstm_ae}, the other input (other than the previous cell state) can be either zero or the output of the previous LSTM cell. The outputs of LSTM layers are fed as input to a series of 1D de-convolutional layers which perform reverse of convolution (transposed convolution) to generate data with same shape as that of input data to encoder.

\begin{figure}
\begin{center}
\includegraphics[width=3.4in]{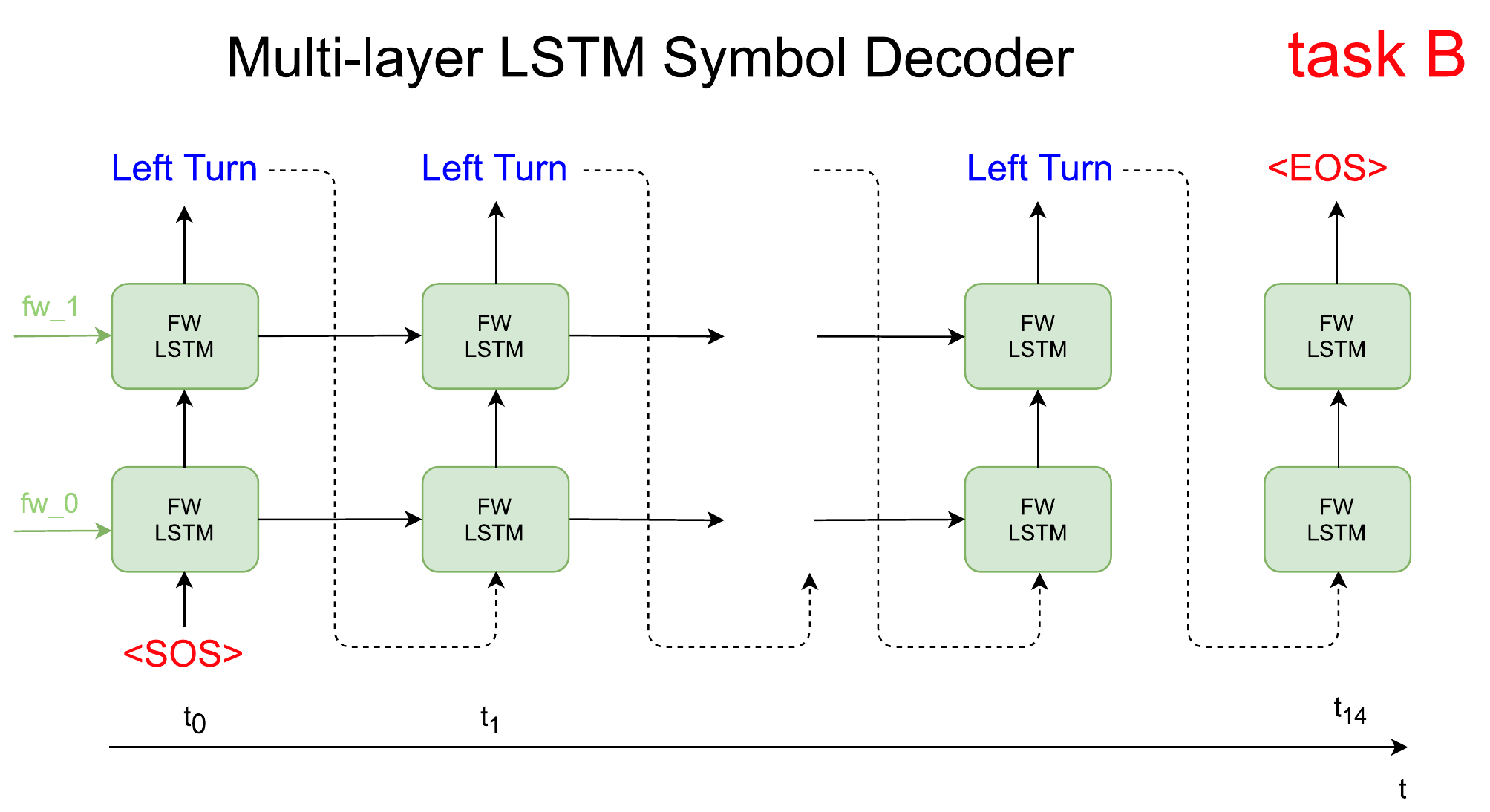}
\end{center}
\caption{Greedy symbol decoder for task~B (maneuver predictor) in Fig.~\ref{fig:lstm_multitask_highlevel}.}\label{fig:lstm_multitask_decoder_taskB}
\end{figure}

\textbf{Decoder (Predictor, Task~B)}: The decoder of the symbol predictor (task~B) is shown in Fig.~\ref{fig:lstm_multitask_decoder_taskB}. It takes only forward cell states from encoder as it has only uni-directional LSTM layers. It adopts a greedy decoder, where the most probable symbol output of the previous LSTM cell is fed as input to the next LSTM cell. The first LSTM cell takes a special symbol <SOS>, denoting start of sequence, as input. Likewise, the last LSTM cell generates <EOS> symbol, denoting end of sequence. The output symbols of all the LSTM cells is the predicted series of next maneuvers (Left Turn, Left Turn, ... Left Turn in Fig.~\ref{fig:lstm_multitask_decoder_taskB}).

\textbf{Training (Step~1)}: The loss function for Task~A is the Mean Square Error~(MSE) between the input data to encoder and the output of decoder. The loss function for Task~B is weighted cross-entropy loss with weights being the inverse of the frequency of maneuvers in the train data. That is, the weight for symbol, $s$ = $w_s = 1/f_s^k$, where $f_s$ is the frequency ratio of maneuver $s$ in the train data and $k$ is determined empirically for best results. The overall network is trained by minimizing the weighted losses of task~A, task~B and regularization losses (i.e., overall loss, $L_O = w_AL_A + w_BL_B + w_RL_R$, where $w_A$, $w_B$
, $w_R$ are the weights for $L_A$, $L_B$, $L_R$, the task~A, task~B and regularization loss respectively.

\begin{figure}
\begin{center}
\includegraphics[width=3.4in]{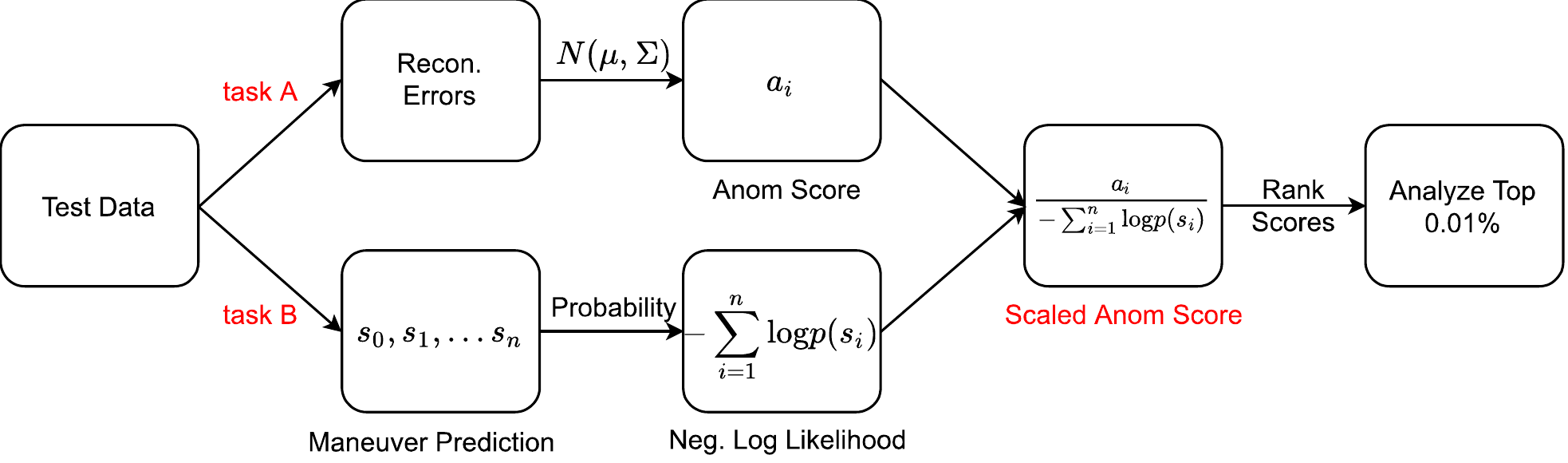}
\end{center}
\caption{Scaled anomaly scores that leverages the maneuver predictions of task~B to reduce number of false positives compared to the scores in Fig.~\ref{fig:anom_scores_combined}.}\label{fig:scaled_anom_scores_test}
\end{figure}

\begin{figure*}[ht!]
        \centering
        \hspace{-0.3in}
           \begin{subfigure}[b]{0.32\textwidth}
        		\centering
        		\includegraphics[width=1\textwidth,height=1.8in]{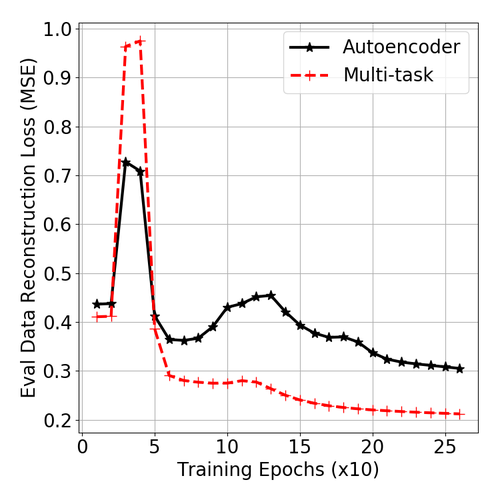}
        		\caption{}
        		\label{fig:eval_epoch_recons_losses}
        	\end{subfigure}
~
        \begin{subfigure}[b]{0.32\textwidth}  
            \centering 
            \includegraphics[width=1\textwidth,height=1.8in]{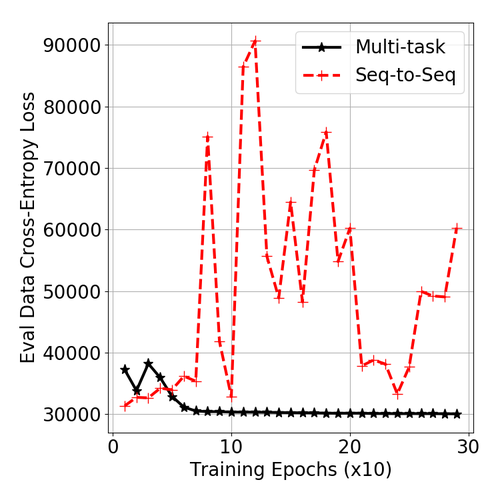}
            \caption{}
            \label{fig:eval_epoch_xent_losses}
        \end{subfigure}
~
        \begin{subfigure}[b]{0.32\textwidth}   
            \centering 
            \includegraphics[width=1\textwidth,height=1.8in]{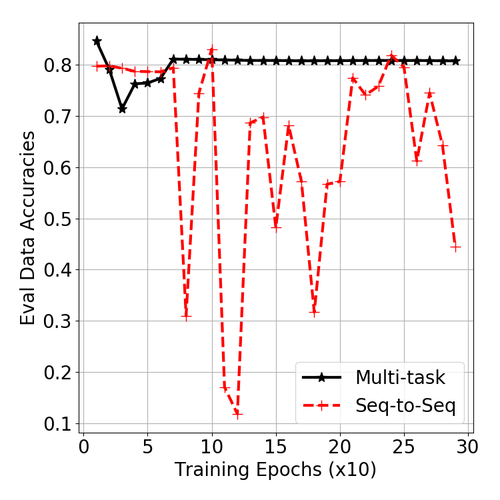}
            \caption{}
            \label{fig:eval_epoch_accuracies}
        \end{subfigure}
        \vspace{-0.1in}
        \caption{\label{fig:recon_results} Comparison of performance on test data between multi-task learning (our approach) and standalone autoencoder or symbol predictor: (a) Reconstruction MSE loss; (b) Cross-entropy loss; (c) Symbol prediction accuracy [\%].}
        \vspace{-0.15in}
\end{figure*}

\textbf{Inference (Step~2)}: During inference, given a test data point, an anomaly score is calculated as mentioned in Sect.~\ref{sec:prop_soln:lstm_ae} and Fig.~\ref{fig:anom_scores_combined}. This anomaly score (say $a_i$), however fares poorly with rare positive classes leading to multiple false positives. In order to address this problem, we define a new anomaly score leveraging the predicted maneuvers from task~B as shown in Fig.~\ref{fig:scaled_anom_scores_test}. Assume $s_0, s_1,...s_n$ correspond to the maneuvers predicted by task~B. We then calculate the negative log-likelihood of such a sequence using $-\sum_{i=1}^n logp(s_i)$ (we assume independence for simplicity). This value is low for more frequent maneuvers (e.g., going-straight) and high for rare maneuvers (e.g., u-turns). We divide $a_i$ with this value to obtain the scaled anomaly score. This is high for more-frequent maneuvers and low for less-frequent maneuvers such as u-turns. In this way, rare but non-anomalous situations are weighed down leading to lesser false positives.

\section{Performance Evaluation}\label{sec:perf_eval}
In this section, we first explain the experimental setup (data and training) then present quantitative and qualitative results for two scenarios---comparison with unsupervised LSTM autoencoder (without using the information of maneuver labels) and semi-supervised multi-class LSTM autoencoder (that uses the information of maneuver labels).

\textbf{Dataset description:} We evaluated our approach on a 150 hours HDD driving dataset~\cite{ramanishka2018CVPR}, which is collected from February 2017 to March 2018, predominantly during day-time. The data consists of Controller Area Network~(CAN) bus data that has information about six driving modalities---steer angle, steer speed, speed, yaw, pedal angle and pedal pressure. The data has been downsampled to $5~\rm{Hz}$ from the original $100~\rm{Hz}$ as we observed better results with lower sampled data. Since this is time-series data, we adopted a sliding-window approach as follows. For both autoencoder and symbol predictor, the size of the input window is $5~\rm{secs}$, with a stride length of $0.5~\rm{secs}$. For symbol predictor, the size of prediction window is $3~\rm{secs}$. In order to obtain meaningful results (e.g., anomalous results corresponding to when the car is parked are not useful), we filtered out those windows where the maximum speed of the vehicle is less than $15~\rm{mph}$. This results in a total of $762671$ datapoints (windows). We then scaled this data between $0$ and $1$ in order to make the network invariant to scales of data. Of this data, 70\% is used for training the models and rest for evaluating the performance (i.e., $533869$ windows for train and $228802$ windows for test). Table~\ref{table:hdd} shows the annotated maneuvers/labels present in the HDD dataset (`Background' indicates going-straight) with corresponding percentage of occurrence.

\begin{table}[t]
\footnotesize
\caption{Distribution of maneuvers/labels in the HDD dataset~\cite{ramanishka2018CVPR}.} \label{table:hdd}
\vspace{-0.1in}
\begin{center}
\setlength\tabcolsep{1.5pt} %
\begin{tabular}{|p{3cm}|p{1.7cm}|}
\hline
\textbf{Label}    & \textbf{Percent [\%]}    \\ \hline \hline
Background    & 87.15 \\ \hline
Intersection Passing   & 6.00 \\ \hline
Left turn    & 2.58 \\ \hline
Right turn   & 2.31 \\ \hline
Left lane change    & 0.54 \\ \hline
Right lane change   & 0.50 \\ \hline
Crosswalk passing    & 0.27 \\ \hline
U-turn   & 0.23 \\ \hline
Left lane branch    & 0.20 \\ \hline
Right lane branch   & 0.08 \\ \hline
Merge    & 0.14 \\ \hline
\end{tabular}
\vspace{-0.2in}
\end{center}
\end{table}

\textbf{Training:} We used tensorflow to build, train and test the models with a minibatch size of 512 windows. Weights for reconstruction loss (task~A), cross-entropy loss (task~B) and regularization loss have been set empirically as follows---$w_A = 1$, $w_B = 0.001$ and $w_R = 0.0001$. We used $k=0.5$ to scale the weights in cross-entropy loss as mentioned in Sect.~\ref{sec:prop_soln:multi_task}. We used two-layers of bi-LSTMs with a hidden size $256$ units for each LSTM cell. We trained the overall network for about $300$ epochs using Adam optimizer~\cite{Kingma2015Adam:Optimization} with a learning rate of $0.01$ and epsilon value of $0.01$. 

\textbf{Comparison with LSTM autoencoder:} We compare our approach with fully unsupervised LSTM autoencoder. The network architecture, training method and parameters are similar to that of the LSTM autoencoder part of our multi-task network.

\begin{figure*}[ht!]
        \centering
        \hspace{-0.3in}
           \begin{subfigure}[b]{0.32\textwidth}
        		\centering
        		\includegraphics[width=1\textwidth,height=1.8in]{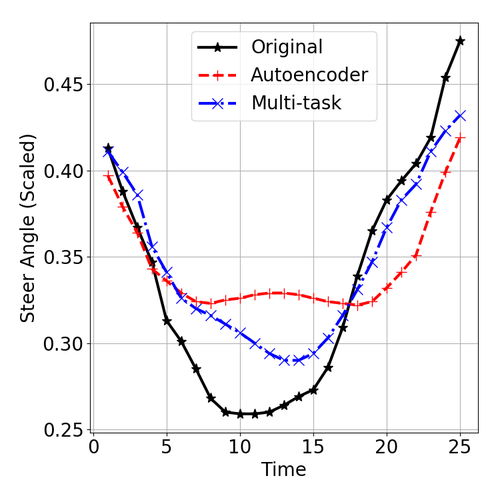}
        		\caption{}
        		\label{fig:left}
        	\end{subfigure}
~
        \begin{subfigure}[b]{0.32\textwidth}  
            \centering 
            \includegraphics[width=1\textwidth,height=1.8in]{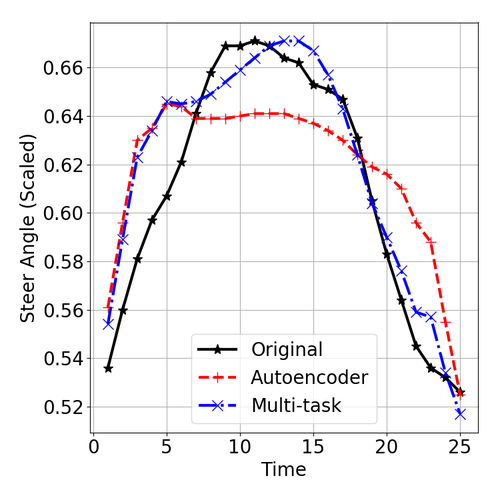}
            \caption{}
            \label{fig:right}
        \end{subfigure}
~
        \begin{subfigure}[b]{0.32\textwidth}   
            \centering 
            \includegraphics[width=1\textwidth,height=1.8in]{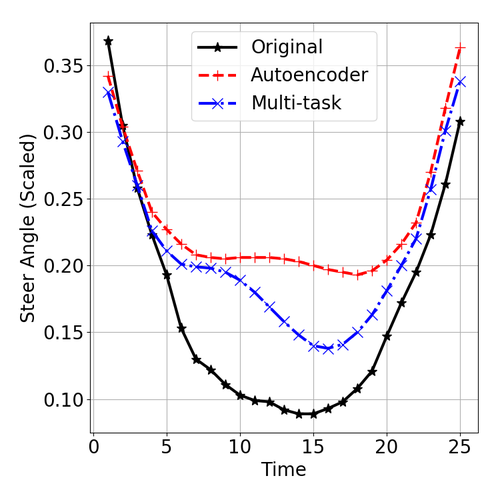}
            \caption{}
            \label{fig:u}
        \end{subfigure}
        \vspace{-0.1in}
        \caption{\label{fig:turn_results} Reconstruction performance of turn data between multi-task learning (ours) and standalone autoencoder: (a) Left turn; (b) Right turn; (c) U-turn.}
        \vspace{-0.15in}
\end{figure*}

\begin{table}[t]
\footnotesize
\caption{Comparison of normalized reconstruction MSE losses.} \label{table:recon_loss}
\vspace{-0.1in}
\begin{center}
\setlength\tabcolsep{1.5pt} %
\begin{tabular}{|p{1.7cm}|p{3cm}|p{2cm}|}
\hline
\textbf{Feature}    & \textbf{LSTM Autoencoder~\cite{Malhotra2016}}    & \textbf{Our Approach}   \\ \hline \hline
Steer Angle    & 0.0005 & 0.0003 \\ \hline
Steer Speed    & 0.0004 & 0.0003 \\ \hline
Speed          & 0.0004 & 0.0003 \\ \hline
Yaw            & 0.0004 & 0.0003 \\ \hline
Pedal Angle    & 0.0012 & 0.0012 \\ \hline
Pedal Pressure & 0.0012 & 0.0003 \\ \hline
Combined       & 0.3043 & 0.2082 \\ \hline
\end{tabular}
\vspace{-0.2in}
\end{center}
\end{table}

\begin{table}[t]
\footnotesize
\caption{Comparison of qualitative results by analyzing top 0.01\% scores.} \label{table:qual_results}
\vspace{-0.1in}
\begin{center}
\setlength\tabcolsep{1.5pt} %
\begin{tabular}{|p{2cm}|p{1.7cm}|p{1.7cm}|p{1.9cm}|}
\hline
\textbf{Category}    & \textbf{LSTM Autoencoder~\cite{Malhotra2016}}    & \textbf{Our Approach} & \textbf{Our Approach (Scaled Scores)}  \\ \hline \hline
Speed       & 21.7\%     & 21.7\%     & 30.4\%     \\ \hline
K-turns     & 13.0\%     & 8.8\%      & 17.4\%     \\ \hline
U-turns     & 4.4\%      & -        & -        \\ \hline
Lane Change & 34.8\%     & 47.8\%     & 39.1\%     \\ \hline
Normal      & 26.1\%     & 21.7\%     & 13.1\%     \\ \hline
Total       & 100\% (23) & 100\% (23) & 100\% (23) \\ \hline
\end{tabular}
\vspace{-0.2in}
\end{center}
\end{table}

\underline{Quantitative results:}  
After the network has been trained, we tested it on evaluation/test data.  Fig.~\ref{fig:eval_epoch_recons_losses} compares the reconstruction MSE loss between our approach and LSTM autoencoder vs. the number of epochs on test data. We can notice that our approach converges to a lower loss. Table~\ref{table:recon_loss} shows the average normalized reconstruction loss on test data for different modalities between our approach (multi-task learning) and LSTM autoencoder. We can notice that, our approach results in lower reconstruction loss with $33\%$ lower error ($0.2$) compared to the standalone autoencoder ($0.3$) in the `combined' category. This shows that the combined system does a better job of learning representations than the standalone autoencoder, resulting in lower loss. Fig.~\ref{fig:eval_epoch_xent_losses} compares the weighted cross-entropy loss between our approach and standalone symbol predictor. We can notice that our approach achieves lower loss than symbol predictor. Also, we can observe that by coupling an autoencoder to a symbol predictor, the zig-zag behavior of the latter has been smoothened out. We can observe similar behavior in Fig.~\ref{fig:eval_epoch_accuracies} for symbol prediction accuracy (as our data is annotated with maneuvers, we are able to calculate the maneuver prediction accuracy with respect to ground truth). Fig.~\ref{fig:turn_results} compares the reconstruction performance of three sample turns in test data---Left, Right, U---between our approach and standalone autoencoder. We can notice in all three cases that our approach does a better job of reconstruction when compared to original data (to get these results, we used scaled steer angle data).

\begin{table}[t]
\footnotesize
\caption{Comparison of normalized reconstruction MSE losses (without u-turn data).} \label{table:recon_loss_2}
\vspace{-0.1in}
\begin{center}
\setlength\tabcolsep{1.5pt} %
\begin{tabular}{|p{1.7cm}|p{4cm}|p{2cm}|}
\hline
\textbf{Feature}    & \textbf{Multi-class LSTM Autoencoder}    & \textbf{Our Approach}   \\ \hline \hline
Steer Angle    & 0.0007 & 0.0004 \\ \hline
Steer Speed    & 0.0005 & 0.0004 \\ \hline
Speed          & 0.0006 & 0.0004 \\ \hline
Yaw            & 0.0006 & 0.0003 \\ \hline
Pedal Angle    & 0.0014 & 0.0013 \\ \hline
Pedal Pressure & 0.0012 & 0.0004 \\ \hline
Combined       & 0.4058 & 0.2456 \\ \hline
\end{tabular}
\vspace{-0.2in}
\end{center}
\end{table}

\begin{table}[t]
\footnotesize
\caption{Comparison of percentage of u-turns detected (qualitative results) by analyzing top anomaly scores.} \label{table:qual_results_2}
\vspace{-0.1in}
\begin{center}
\setlength\tabcolsep{1.5pt} %
\begin{tabular}{|p{1.3cm}|p{2.3cm}|p{2.2cm}|p{2.3cm}|}
\hline
\textbf{Percentile Top Scores}    & \textbf{Multi-class LSTM Autoencoder}    & \textbf{Our Approach} & \textbf{Our Approach (Scaled Scores)}  \\ \hline \hline
0.001       & 0.39\% (3/765)     & 1.70\% (13/765)      & 7.97\% (61/765)     \\ \hline
0.01     & 1.96\% (15/765)      & 7.97\% (61/765)       & 29.02\% (222/765)     \\ \hline
0.1     & 13.33\% (102/765)       & 17.25\% (132/765)         & 48.63\% (372/765)       \\ \hline
0.5    & 73.46\% (562/765)      & 52.68\% (403/765)     & 84.44\% (646/765)     \\ \hline
1      & 100.00\% (765/765)     & 99.87\% (764/765)      & 100.00\% (765/765)     \\ \hline
\end{tabular}
\vspace{-0.2in}
\end{center}
\end{table}

\underline{Qualitative results:}
After the network is trained, the reconstruction errors (of dimension $25$ due to $5~\rm{Hz}$ sampling for $5~\rm{secs}$) for each modality are fit to $25$-variable gaussian distribution as explained in Sect.~\ref{sec:prop_soln:multi_task}. We also considered another modality which is a combination of all of them. The errors corresponding to this combined modality are fit to a $300$-variable ($25 \times 6$) gaussian distribution. We then passed the test data (in windows) to the network and calculated the mahalanobis distances (anomaly scores) for each window of data as per Fig.~\ref{fig:anom_scores_combined}. We also calculated the scaled anomaly scores using the predicted maneuvers by dividing the anomaly scores with the negative log-likelihood of the predicted maneuvers as per Fig.~\ref{fig:scaled_anom_scores_test}. For both cases (scaled and non-scaled), we analyzed the top $0.01\%$ scores and their corresponding windows. For this purpose, we extracted the video segment corresponding to each window and manually inspected to check if there is any anomalous behavior. By analyzing the video segments corresponding to top $0.01\%$ anomaly scores, we could classify them into five categories---`Speed' anomalies (e.g., abrupt braking), `K-turns', `U-turns', `Unusual lane change' and finally `Normal' (no anomaly has been noticed when inspected visually). We have summarized our analysis results in Table~\ref{table:qual_results}. We can notice that, while autoencoder classifies U-turns as anomalous, our approach (both scaled and unscaled) does not. We can also notice that our scaled approach classifies lesser `Normal' and more `Speed' anomalies. By comparing, the percentage of `Normal' cases classified as anomalous, we can tell that scaled approach performs better than unscaled, which in turns performs better than standalone autoencoder approach. We have included a video demo showing the different kinds of anomalies (listed in Table~\ref{table:qual_results}) detected using the above approaches along with the submission.

\begin{figure}
\begin{center}
\includegraphics[width=3.4in,height=3in]{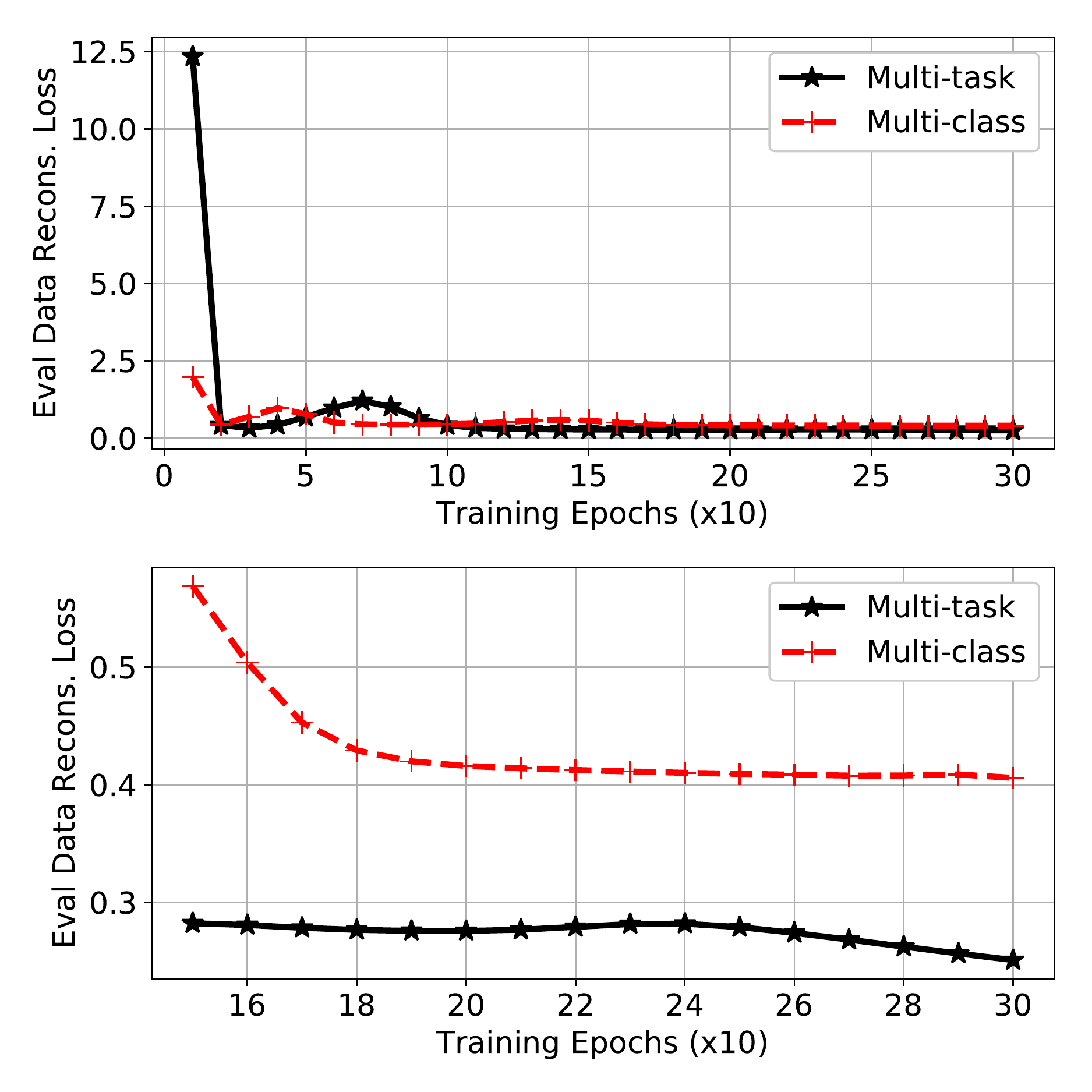}
\end{center}
\caption{(Top) Comparison of eval data MSE reconstruction loss between Multi-task (our approach) and multi-class LSTM autoencoder. (Bottom) Zoomed version of above showing MSE reconstruction loss for 150 to 300 training epochs, for clear visualization. }\label{fig:multi_class_eval_recons_losses}
\end{figure}

\balance

\textbf{Comparison with multi-class/ensemble LSTM autoencoder: }While the above fully unsupervised LSTM autoencoder did not make use of the maneuver labels, we compared our approach with multi-class LSTM autoencoder that makes use of the maneuver labels like our approach. For this purpose and in order to test the performance of the algorithms, we considered one of the maneuvers viz., u-turn as an anomaly. That is, after we split the entire data into train and test data windows, we discarded those windows in the train data where the majority maneuver is a u-turn. The remaining train data, which is mainly devoid of any u-turn windows, is fed to our multi-task classifier. For multi-class LSTM autoencoder, we further divided this train data into 10 parts, each part corresponding to one of the 10 maneuvers in Table~\ref{table:hdd} except U-turn. Then we trained 10 LSTM autoencoder classifiers (i.e., an ensemble) corresponding to these 10 maneuvers by providing only the data specific to that maneuver. When given a test data/window, each of the 10 classifiers are used to find 10 reconstruction loss values. Then the lowest of these is considered the reconstruction loss for that test data point.

\underline{Quantitative results.} Fig.~\ref{fig:multi_class_eval_recons_losses} shows the quantitative results, which compare the eval data MSE reconstruction loss between our approach (Multi-task) and multi-class LSTM autoencoder approach as the number of training epochs is increased. We recall that the reconstruction loss for multi-class approach is obtained as the lowest reconstruction loss corresponding to 10 different class (maneuver)-specific autoencoder classifiers. We can observe that Multi-task approach finally achieves a lower loss compared to multi-class approach. The final (after 300 epochs of training) reconstruction loss on the eval/test data for each feature is summarized in Table~\ref{table:recon_loss_2}. We can notice that our approach achieves lower reconstruction error for all features, compared to multi-class approach.

\underline{Qualitative results.}
In order to evaluate the qualitative performance of the algorithms, we first sorted the reconstruction losses/scores in decreasing order and then found the number of u-turn windows detected in the test data by each approach in the top $0.001,0.01,0.1,0.5,1$ percentile anomaly scores. The results are shown in Table~\ref{table:qual_results_2}. For our multi-task approach, we have two scenarios---actual reconstruction loss and scaled reconstruction loss. Considering the especially the top percentiles, we can notice that our approach with scaled scores performs better than our approach with normal scores which in turn performs better than multi-class approach. For example, considering the top $0.001$ percentile anomaly scores for each approach---our approach with scaled scores is able to detect $7.97\%$ i.e., $61$ of a total $765$ u-turn windows in test data (consisting of $228802$ windows), while this number is $1.7\%$ for our approach with actual scores and only $0.39\%$ for multi-class autoencoder approach.

\section{Conclusion and Future Work}\label{sec:conc}
We have presented a multi-task learning based anomaly detection framework that performs better than existing LSTM autoencoder based appraoches. We leverage domain knowledge to reduce false positives. We have validated the proposed approach on 150 hours of driving data and showed the benefits of our approach both quantitatively and qualitatively. \\
Though we have seen some artifacts in the data corresponding to `Normal' cases (leading them to be classified as anomalous), we will investigate why some other `Normal' cases are classified as anomalous. This along with, improved network architectures and the use of video data (not just CANbus) will be the focus of our future work.

\bibliographystyle{IEEEtran}%
\bibliography{references}

\end{document}